\documentclass{article}

\usepackage{arxiv}
\usepackage{amsmath}

\usepackage[utf8]{inputenc} 
\usepackage[T1]{fontenc}    
\usepackage{hyperref}       
\usepackage{url}            
\usepackage{booktabs}       
\usepackage{amsfonts}       
\usepackage{nicefrac}       
\usepackage{microtype}      
\usepackage{cleveref}       
\usepackage{lipsum}         
\usepackage{graphicx}
\usepackage[square,sort,comma,numbers]{natbib}
\usepackage{doi}
\usepackage{textcomp}
\usepackage{booktabs}
\usepackage{subcaption}
\usepackage{xcolor}
\usepackage{tabularx}
\usepackage{tabularray}
\newcolumntype{Y}{>{\centering\arraybackslash}X}
\newcommand\blfootnote[1]{%
  \begingroup
    \renewcommand\thefootnote{}\footnote{#1}%
    \addtocounter{footnote}{-1}%
  \endgroup
}

\title{Unsupervised Deep Learning-based Keypoint Localization Estimating Descriptor Matching Performance}

\newif\ifuniqueAffiliation
\uniqueAffiliationtrue

\ifuniqueAffiliation 

\usepackage{authblk}

\newbox{\orcid}\sbox{\orcid}{\includegraphics[scale=0.06]{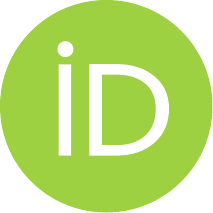}} 
\author[1,2]{%
	\href{https://orcid.org/0000-0001-7824-8098}{\usebox{\orcid}\hspace{1mm}David Rivas-Villar\thanks{\texttt{Corresponding author: david.rivas.villar@udc.es}}}%
}
\author[1,2]{%
	\href{https://orcid.org/0000-0002-9080-9836}{\usebox{\orcid}\hspace{1mm}Álvaro S. Hervella\thanks{\texttt{a.suarezh@udc.es}}}%
}
\author[1,2]{%
	\href{https://orcid.org/0000-0003-4407-9091}{\usebox{\orcid}\hspace{1mm}José Rouco \thanks{\texttt{jrouco@udc.es}}}%
}
\author[1,2]{%
	\href{https://orcid.org/0000-0002-0125-3064}{\usebox{\orcid}\hspace{1mm}Jorge Novo\thanks{\texttt{jnovo@udc.es}}}%
}
\affil[1]{Grupo VARPA, Instituto de Investigacion Biomédica de A Coru\~na (INIBIC), Universidade da Coru\~na, 15006 A Coru\~na, Spain}
\affil[2]{Departamento de Ciencias de la Computación y Tecnologías de la Información, Universidade da Coru\~na, A Coruña, 15071, A Coruña, Spain}
\fi

\renewcommand{\shorttitle}{Unsupervised Keypoint Localization Estimating Descriptor Matching Performance}

\hypersetup{
pdftitle={Unsupervised Deep Learning-based Keypoint Localization Estimating Descriptor Matching Performance},
pdfsubject={cs.CV},
pdfauthor={David~Rivas-Villar, Álvaro S.~Hervella, José~Rouco, Jorge~Novo},
pdfkeywords={ Retinal Image Registration, Unsupervised Learning, Color Fundus, Medical Imaging},
}

\begin{document}
\maketitle
\blfootnote{\fontsize{6}{7}\selectfont \copyright~2026 IEEE. Personal use of this material is permitted. Permission from IEEE must be obtained for all other uses, in any current or future media, including reprinting/republishing this material for advertising or promotional purposes, creating new collective works, for resale or redistribution to servers or lists, or reuse of any copyrighted component of this work in other works. Accepted for publication in \emph{IEEE Journal of
Biomedical and Health Informatics}.
DOI: \href{https://doi.org/10.1109/JBHI.2026.3695387}{10.1109/JBHI.2026.3695387}}
\begin{abstract}
Retinal image registration, particularly for color fundus images, is a challenging yet essential task with diverse clinical applications. Existing registration methods for color fundus images typically rely on keypoints and descriptors for alignment; however, a significant limitation is their reliance on labeled data, which is particularly scarce in the medical domain. 

In this work, we present a novel unsupervised registration pipeline that entirely eliminates the need for labeled data. Our approach is based on the principle that locations with distinctive descriptors constitute reliable keypoints. This fully inverts the conventional state-of-the-art approach, conditioning the detector on the descriptor rather than the opposite.

First, we propose an innovative descriptor learning method that operates without keypoint detection or any labels, generating descriptors for arbitrary locations in retinal images. Next, we introduce a novel, label-free keypoint detector network which works by estimating descriptor performance directly from the input image.

We validate our method through a comprehensive evaluation on four hold-out datasets, demonstrating that our unsupervised descriptor outperforms state-of-the-art supervised descriptors and that our unsupervised detector significantly outperforms existing unsupervised detection methods. Finally, our full registration pipeline achieves performance comparable to the leading supervised methods, while not employing any labeled data. Additionally, the label-free nature and design of our method enable direct adaptation to other domains and modalities.

\end{abstract}


\section{Introduction}
\label{sec:introduction}

Retinal Image Registration (RIR) is the process of aligning retinal images based on their visual content. Typically, registration approaches in RIR are used either to create a wider field of view by stitching together narrower images, or to align images captured at different time points, enabling longitudinal studies. Thus, RIR is a highly important task, supporting multiple clinical applications \cite{survey}. Nonetheless, manual alignment is not possible in busy, day-to-day clinical workflows. Therefore, the development of efficient and robust automated registration methods is desirable. In retinal images, structures such as the blood vessels and the optic disc contain the relevant patterns for registration. On the contrary, the extensive homogeneous background is, generally, not relevant for registration \cite{rivas}.

Within RIR, Color Fundus (CF) images are of particular relevance. This imaging technique is very widespread and a very cost-effective \cite{costeffec,Besenczi} tool for the diagnosis of ophthalmic \cite{Kanski} and even systemic diseases \cite{liu2023retinal}. However, besides the general characteristics of retinal images, CF images present certain particularities that complicate the registration process. CF images can have multiple imperfections. Some examples include, but are not limited to, blurriness, underexposure, overexposure, glares from light reflections, motion artifacts from shifting gaze, highlights or reflections due to incorrect distances between camera and subject, etc. These technical imperfections coupled with the morphological changes in the retina, caused by progression or remission of diseases or the appearance of lesions, make CF registration a highly challenging process \cite{rivas, eccv20, rempe}.

In medical imaging, classical approaches (i.e., non-deep learning) have traditionally yielded superior results. However, deep learning has gained preference due to its capacity to learn directly from data, eliminating the need for manual tuning or feature engineering required by classical approaches. Beyond being time-consuming, manual tuning in classical approaches also makes it harder to adapt models to new domains. Whether registration methods are classical or deep learning-based, they are classified into three groups: Intensity-Based Registration (IBR), Direct Parameter Regression (DPR), and Feature-Based Registration (FBR) \cite{rivas,voxel}.

Generally, in medical image registration most current state-of-the-art methods are based on IBR or DPR \cite{voxel,dlir,Haonan}. On the one hand, IBR methods iteratively maximize a similarity metric among the images by modifying the transformation parameter space. On the other hand, DPR methods directly predict deformation fields for the images using deep neural networks \cite{Haskins,voxel}. The output of these networks is a deformation field which is applied to the moving image. Then, the transformed moving image is compared with the fixed one using a similarity metric, which acts as the loss function.

FBR methods rely on keypoints, which are distinctive spatial locations that can be detected in the images to be registered. Keypoints can either be generic, meaning they apply to any type of image, or domain specific, meaning that they are suited only to a particular type of image. These keypoints drive the registration process. Additionally, since they are easily displayable, FBR methods are inherently explainable, which is an advantage over competing approaches. Furthermore, to aid in the transformation estimation, descriptors are calculated for each keypoint. Descriptors are feature representations that, ideally, uniquely characterize each keypoint, allowing for fast and easy matching among keypoints.

The specifics of CF images like small and scattered registration-relevant features (blood vessels, optic disc), low overlap and large rigid transformations, image artifacts, etc. make both IBR and DPR approaches generally unsuitable for CF registration \cite{rivas3}. Therefore, due to the particularities of CF images, registration methods for these images are limited to the FBR paradigm \cite{rivas3, eccv20}. The best performing deep learning-based methods for CF are either supervised or detector-less methods. Supervised detector-based methods rely on annotated ground truth for domain-specific keypoints, while detector-less methods are more computationally complex because they hierarchically match descriptors across the entire image instead of just a selected set, belonging to the detected keypoints. Therefore, we propose a novel unsupervised approach to both train descriptors and learn keypoints from them. Our experimental results demonstrate that our method yields a robust detector and descriptor despite the absence of supervision, capable of matching the performance of state-of-the-art approaches, even accounting for efficiency.

The main contributions of this paper are:
\begin{itemize}
    \item We propose a novel unsupervised descriptor training method, improving the results of supervised alternatives.

    \item Based on the premise that locations with distinctive descriptors constitute reliable keypoints, we propose a novel unsupervised approach to train detector networks.

    \item We propose several alternatives to measure descriptor performance and study the different detectors produced.
    
    \item We analyze our proposal's performance extensively, using four different hold-out datasets and varied evaluations.
    
    \item We show that our registration pipeline achieves comparable performance to state-of-the-art methods, without any labeled data, a significant step in RIR, given the scarcity of labels in medical data.
    
\end{itemize}

\section{Related Work}

In order to evaluate the performance of CF registration, the FIRE dataset \cite{fire} is commonly used as it is the only dataset with a suitable ground truth. This limits the evaluation of CF RIR methods to pairwise image alignment. However, it is important to note that most registration methods require little to no adaptation to handle multiple images. Currently, classical methods are the ones still dominating this domain in terms of performance in this benchmark dataset. The best results in FIRE are obtained by the classical approach VOTUS \cite{votus}, which matches blood vessel graphs and uses the transformation with the most degrees of freedom in the state of the art. REMPE \cite{rempe}, another classical approach, also provides remarkable results by mixing specific and generic keypoints and estimating an eye-specific transformation. However, state-of-the-art registration approaches in CF images are moving away from classical approaches towards the use of deep learning since it has multiple advantages \cite{rivas2}. The strengths of deep learning methods stem from their data-driven learning, which allows these methods to be much more adaptable than classical ones. Deep learning models can automatically learn task-relevant features when provided with a suitable loss function, whereas classical methods typically rely on manually tuned, hand-crafted features and heuristics. Thus, deep learning methods are much more versatile, easily adapting to different datasets, imaging devices or even domains without any laborious human intervention.

Current state-of-the-art deep learning methods can be categorized into detector-based \cite{rivas3, eccv20} and detector-less \cite{geoformer}. Detector-based approaches generally detect and then describe keypoints in order to match them. In contrast, detector-less or detector-free approaches forgo the detection step and, instead, match descriptors directly \cite{geoformer}. First, they do it in a coarse manner and then refine the matching more finely, at higher resolution. Additionally, detector-based methods can be supervised or unsupervised. The advantage of unsupervised methods is that they can be trained in any dataset, device or domain and without any ground truth. Supervised keypoint detectors generally outperform unsupervised ones, highlighting a trade-off between accuracy and flexibility.

The best detector-based method is the supervised SuperRetina with Knowledge Distillation \cite{kdsr} which improves upon SuperRetina \cite{eccv20}. SuperRetina \cite{eccv20} is itself based on a general image registration approach, SuperPoint \cite{superpoint}. SuperRetina uses supervised keypoints and it is able to detect more keypoints than originally included in the ground truth using a novel keypoint expansion module. SuperRetina is subsequently improved slightly in \cite{kdsr} adding reverse knowledge distillation for keypoint location and employing larger, heavier models. Similarly, ConKeD \cite{rivas3} and ConKeD++ \cite{rivas4} are also based on supervised detection, but rely on a novel contrastive multi-positive multi-negative description loss that improves data efficiency. These approaches demonstrate equivalent results to SuperRetina while having significant advantages related to efficiency. On the other hand, the best detector-free method, GeoFormer \cite{geoformer}, builds on LoFTR \cite{loftr}, incorporating RANSAC \cite{ransac} to improve its matching performance on RIR.

Overall, no deep learning method reaches the performance level of classical methods (i.e., VOTUS \cite{votus}) in spite of their adaptability advantages. Similarly, no detector-free method reaches the performance obtained by supervised detector-based methods. Additionally, current  unsupervised detector-based  methods, combining the advantages of detector-based (simplicity and straight-forward matching) and non-supervision from detector-less methods, are not competitive as they require many keypoints and do not offer sufficient performance \cite{rivas2}. Hence, this paper introduces a novel unsupervised, detector-based methodology leveraging a novel keypoint detector capable of estimating the matching performance of the description network.

The idea of detecting keypoints from descriptors  (describe-to-detect) was pioneered in D2\cite{d2} and expanded in R2D2 \cite{r2d2}. Typical detector-based approaches either train descriptors specific to predefined keypoints or jointly optimize keypoints and descriptors. However, in describe-to-detect, the descriptors are actually used to train or create a descriptor. Particularly, in D2 a novel methodology to detect keypoints based on the descriptors is proposed by directly operating with the already calculated descriptor block. In R2D2 both detector and descriptor are trained jointly. This approach creates two separate heatmaps for keypoint detection, one measuring repeatability and other reliability. However, one key disadvantage of this method is that the heatmaps are dependent on several parameters that must be fixed at training time, like a reliability threshold. We propose to create an unsupervised descriptor network, which can work paired with any detector as it is not conditioned to any keypoint.  From it, we bootstrap a separate predictor network that estimates the matching performance of descriptors, directly from the input image. This effectively removes D2's heuristic computation and R2D2's ad-hoc parameters.

\section{Methodology}

The presented approach is based on the principle that locations with accurate and distinctive descriptors constitute reliable keypoints. Consequently, a description network can be used to train a detection network tailored to its descriptors. This represents an inversion of the typical state-of-the-art training pipeline. Other methods, such as ConKeD, use keypoints to condition descriptors. Similarly, SuperRetina jointly trains both keypoints and descriptors, creating a mutually conditioned detection-and-description approach where the detector and descriptor cannot be separated. On the other hand, our approach leverages an unsupervised descriptor network, which is not conditioned to any keypoint detector, to bootstrap its own detector. To achieve this, we explore various approaches for generating heatmaps from the descriptors produced by the description network. The goal of these heatmaps is to estimate the matching performance of the pixel-wise descriptors produced by the trained descriptor network, serving as ground truth for training the detector network. After training, the detector network, which only receives as input the image, is capable of estimating the performance of the descriptor network for each discrete location, making it an effective keypoint detector. As both the descriptor and detector network are trained without any labels, our entire methodology is unsupervised. Importantly, since our method is not bound by domain-specific keypoints, it is directly adaptable to any other image modality.

\subsection{Unsupervised Descriptor Training}

We build upon the state-of-the-art approach ConKeD++ \cite{rivas4}. This framework trains a descriptor network using a contrastive learning approach. In particular, the method follows a multiview strategy employing multiple positive samples as well as multiple negative ones for each anchor point (i.e., the reference keypoint). All the sample points are selected based on the inference of a supervised neural network, which detects domain-specific keypoints (blood vessel crossovers and bifurcations). In this work, we redesign the framework to make it unsupervised while maintaining its multi-view, multi-positive, multi-negative approach. This keeps the high data efficiency while eliminating the dependency on labeled ground truth data. In particular, instead of relying on a supervised detector network, we train the descriptor network using randomly sampled points.  This eliminates any dependency on labeled data and allows to learn descriptors for arbitrary keypoints in the retina. We name this approach \textbf{UnConKeD} (\textbf{Un}supervised \textbf{ConKeD}++). An overview of the training methodology can be seen in Figure \ref{fig:method}-A.

The training is conducted using a batch of multiple images, where they are all different views (i.e., image augmentations) of the same original one. The batch is of size $N+1$, (i.e., $N$ augmentations and $1$ original image). We randomly sample points from the original image and then transform them to match the augmentations of the different views. Each of the points is used as an anchor and it is compared to multiple positives (i.e., the same point augmented in the various image views) and multiple negatives (all of the other points in all the images, including the one from where the anchor is sampled). These comparisons, which are performed among the descriptors, are used to compute the training loss.

In order to train the descriptor network, we choose Fast AP loss \cite{fastap,rivas4} as it provides superior performance in a variety of datasets and tests in RIR \cite{rivas4}. Furthermore, this loss does not require a temperature parameter \cite{rivas4, fastap}, unlike other common contrastive losses \cite{vinyals, supcon}. The temperature could prove complex to tune, especially when varying imaging domains. Furthermore, the implicitly structured embedding space created by the loss \cite{rivas4} could positively impact descriptor performance when learning from randomly sampled points. This loss is defined as:

\begin{equation}
    \mathcal{L}_{FastAP} =\frac{1}{|S|}\sum_{i =  1}^{S}{\frac{1}{M^+_{s_{i}}}\sum_{j= 1}^{Q}\frac{H^{+}_{j}h^{+}_{j}}{H_{j}}},
\label{eq:1}
\end{equation}

where $S$ is the set of samples (i.e., keypoints) in the batch. The cardinality of $S$ is $|S|$. In our case, as we use a multiviewed batch of $N+1$ images, each of them containing $K$ keypoints, the cardinality of $S$ would be $|S|=(N+1)K$. $s_{i}$ represents a particular sample from $S$, identified by its index $i$. This sample acts as the anchor and  $M^{+}_{s_{i}}$ is the number of positives that this sample has in the batch. Following the ConKeD++ approach $M^{+}_{s_{i}} = N$. $Q$ is the number of discrete bins in which the histogram is quantized.  $h_{j}$ is the number of samples that fall in the $j-$th bin in the histogram and $H_{j} = \sum_{k \le j} h_{k}$ is the cumulative sum of the histogram up to that bin. $h^{+}_{j}$ is the number of positive samples of $s_{i}$ in the $j-$th bin of the histogram and $H^{+}_{j}$ is its cumulative sum.

It could be argued that the random sampling used to
train descriptors could limit their performance. Specifically, in CF images, since most of the pixels within the RoI (Region of Interest) belong to the background, many sampled points will consequently fall on areas that are typically not considered relevant for the registration or are difficult to uniquely describe for the network.  However, rather than being a limitation, this represents an advantage of our method. First, it must be noticed that we can precisely control the number of samples per batch as opposed to relying on the number of crossovers and bifurcations present in the image, which varies depending on the image and the patient. Hence, we can increase the number of samples used to compute the loss, ensuring high diversity and enough representation of the areas considered relevant for the registration. Moreover, we can also train with a potentially unlimited number of images as we are not limited by the scarcely available labeled data. This can help to learn better descriptors even for the most challenging areas of the image. Additionally, ConKeD++ is specifically designed to describe the points it was trained on (blood vessel crossovers and bifurcations). Thus, the descriptor network only receives training feedback for these specific keypoints. Therefore, if the detector mistakenly identifies a point outside of crossovers and bifurcations, the descriptor network may fail to describe it accurately, potentially leading to mismatches. On the other hand, our proposal trains with every possible point available in the retina, ensuring that no point would cause issues. Therefore, while supervision can accelerate descriptor training by focusing it on a narrow type of keypoints, it may harm the robustness of the method as it only allows feedback for a very small fraction of all the possible keypoints. In this regard, the increased diversity of training samples in our proposal enhances network robustness and, when coupled with our novel unsupervised detector, can even allow for the discovery of new keypoints in the images.

\subsection{Heatmap regression for keypoint detection}

\begin{figure*}
    \centering
    \includegraphics[width=0.8\textwidth]{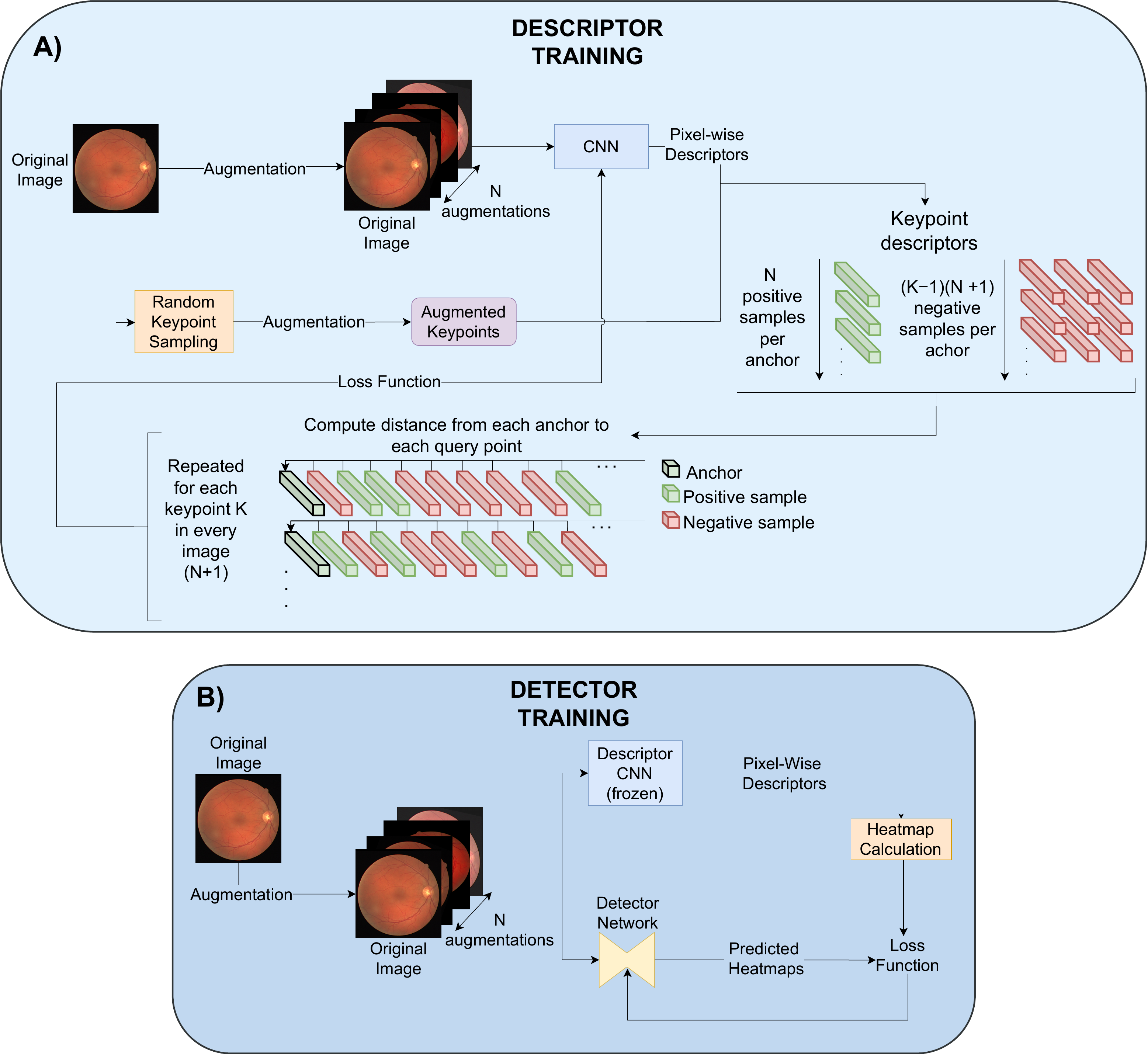}
    \caption{Overview of the proposed training methodology.  A) corresponds to the unsupervised descriptor training,  B) to the unsupervised keypoint detector training.}
    \label{fig:method}
\end{figure*}

Given the trained description network, estimating the matching performance of the descriptors can highlight where reliable keypoints could be located. Inherently, good keypoints should be easily matched by the descriptors. So, conversely, points that are distinctive descriptor-wise should be good keypoints. We propose to train a network that, from the input image, can create a heatmap that is capable of accurately estimating the descriptors' matching performance. Thus, it can create a capable keypoint detector. Importantly, this approach allows the detector to discover relevant keypoints outside of the typically annotated ones. Additionally, since it is not limited or influenced by any labels, it is directly adaptable to any domain, following the unsupervised training of the descriptor network. An overview of the proposed training methodology can be seen in Figure \ref{fig:method}-B.

To train the detector network, first, an input batch of images is processed by the trained descriptor network with frozen weights (i.e., not training). The input batch is like the one used in the descriptor network, a set of views of an original image. This creates a set of dense descriptors for different views of an image. We propose two separate alternatives for estimating the performance of the described keypoints and, thus, creating the target heatmaps for the detection network.

    \textbf{AP Loss}:  
    A straightforward approach to estimate the descriptor matching performance is to estimate the AP loss of each point. Firstly, it is the loss used to train the descriptor network, thus, its use is direct and intuitive. Secondly, any keypoint that produces low AP loss values is inherently reliable as it can be distinguished from many others in the image. In other words, the better the keypoint the lower the loss it produces, meaning that a low AP loss point is discriminative. Therefore, by comparing the descriptors against other samples, like during the descriptor network training, we can create a map of the AP loss values for each point in the image. This map is computed using the loss function defined in Equation \ref{eq:1}, independently computed for each anchor point.

    \textbf{Self-Similarity (SS)}: This represents the similarity between each point and its corresponding positive samples in the rest of the augmented views. Similarly to AP loss, this inherently codifies descriptor performance and, therefore, how reliable a keypoint is. Keypoints very similar to themselves, even under heavy augmentations, are inherently discriminative. In order to compute the Self-Similarity map, we calculate the cosine similarity between each particular point and its correspondences. Next, for each point, we average the similarities corresponding to each augmented view, creating a single map. This way, for any point in the original image, the SS is computed as:

\begin{equation}
    {SS} =\frac{1}{|P|}\sum_{\substack{i =  1}}^{P}Cosim(P^a_{i},P^b_{i}),
\label{eq:2}
\end{equation}

where $P$ is the set of pairwise combinations of descriptors for all the positive samples in the batch (including the reference point in the original image). The cardinality of this set, $|P|$, is equal to $C(M,r)$, where $M$ is the number of views that make up the batch and $r=2$ is the number of elements selected at once. Meanwhile, $a$ and $b$ represent the indexes corresponding to the first and second descriptors in each pair, respectively. $Cosim$ represents cosine similarity. This way, as opposed to the AP heatmap, in the SS heatmap higher is better, as it measures cosine similarity. It should be noted that the choice of image augmentations used to generate different views in the batch may cause some keypoints to vanish in specific views. As a result, the number of pairwise combinations for a given keypoint is given by $|P| = C(M',r)$,  where $M'$ denotes the number of views in which that keypoint is visible. The value of $M'$ may vary, but it always satisfies $1 \leq M' \leq M$. Note that, if $M' = 1$ no loss calculation for that particular keypoint would be possible. However, that case is only reserved for very extreme augmentation.

Both maps are dependent on the variations derived from the augmentations in relation to the positive samples. However, Self-Similarity has one key advantage over AP loss since it does not depend on negative samples. Using negative samples introduces an additional source of randomness and variability, which can make the resulting heatmap more inconsistent and less precise due to the stochastic nature of random keypoint sampling. 
Compared to AP, SS lacks negative contrasts, which may limit its power to detect repeatability problems. This would primarily affect repetitive patterns that differ from the rest of the image but are indistinguishable from themselves.

Once the heatmap is computed, it acts as ground truth for the detection network which is trained to predict it. To train this network the loss function is computed between the predicted heatmap and the one produced either with AP or SS.
Additionally, we study the combination of both maps, as they focus on different properties of the descriptors, to test whether joining them is beneficial. To do so, we multiply both predicted maps inverting the AP one, as the maps represent matching performance in opposite ways (i.e., lower is better in AP, higher is better in SS). This creates the combined heatmap that predicts performance by combining both approaches.

In any case, we use the Mean Square Error (MSE) between the predicted and target heatmaps as the loss function, which for this task can be defined as:

\begin{equation}
{MSE}(hm, {hm}') = \frac{\sum_{i=0}^{P - 1} ({hm}_i - {hm}'_i)^2}{P}
,
\label{eq:3}
\end{equation}

where $hm$ is the ground-truth heatmap (whether it was generated with AP, SS or their combination) and $hm'$ is the one predicted by the network. $P$ is the number of images contained in the batch.

\subsection{Matching Approach}

At inference time, after training both the detector and the descriptor network, they are run at the training image size of $565\times565$. In order to extract keypoints from the heatmaps produced by the detector network, we use non-maximum suppression on the heatmaps, using a window-size of $11\times11$ pixels. Then, we can select the desired amount of keypoints from those available on the heatmap based on their value (lower is better in AP, higher is better in SS). Next, the descriptors for the selected keypoints are obtained from the output of the descriptor network. These descriptors serve to match their corresponding keypoints. In particular, descriptors are matched based on cosine distance. Following the approach used in previous works \cite{rivas3, rivas4}, a match is counted only if the pair of descriptors are mutual nearest neighbors. That is, for two descriptors \textit{A} and \textit{B}, a match is established if \textit{A} is the closest to \textit{B}, and \textit{B} is also the closest to \textit{A}. After this step, the keypoints whose descriptors match are scaled to the source image resolution and the registration pipeline is finalized by applying RANSAC \cite{ransac} to obtain a homographic transformation, like most current methods \cite{rivas4, eccv20}. 

 \section{Experimental Setup and Evaluation}
\subsection{Training Setup}

Regarding the training of the descriptor network, to directly compare our descriptors against those produced by the reference state-of-the-art work ConKeD++ \cite{rivas4}, we use the same network and training setup. In particular, we use a modified L2-Net \cite{l2net,r2d2}, the Adam optimizer \cite{adam} with learning rate $10^{-4}$, and $Q=10$ bins for the AP Loss. We use a batch size of 10 images (1 original and 9 augmentations). The network is trained from scratch for 1000 epochs using the Messidor-2 dataset \cite{mess,mess2}. This dataset is composed of 1748 images captured with 45º of FOV (field of view) from multiple diabetic retinopathy examinations. The images present varying sizes, particularly $1440\times960$, $2240\times1488$ and $2304\times1536$ pixels, however, they are all normalized to $565\times565$ as it is the input image size used in \cite{rivas4}. However, it should be noted that any input image size may be used.

We use the same augmentation approach as the reference work \cite{rivas4}, both for spatial and color augmentations. In terms of spatial augmentations, we use random affine transformations with rotations of $\pm 60^{\circ}$, translations of $0.25\times imageSize$ in each axis, scaling between $(0.75-1.25)\times imageSize$ and shearing of $\pm 30^{\circ}$. For color augmentation, we use random changes in the HSV color space \cite{hsv} as well as random Gaussian Noise with a mean of 0 and a standard deviation of 0.05 and with a probability of being applied of 0.25.
From each image, 1460 keypoints are randomly sampled to compute the AP Loss as these are the most allowed by our hardware.

In terms of the detector network training, the target heatmaps (either AP or SS) are generated using the same configuration that was used for the training of the descriptor network (i.e., same batch size and augmentation regime). The detector network is a slightly modified U-Net \cite{ronneberger15} with 3 input channels (i.e., color image) and half the number of internal channels (i.e., 32 base channels). 
In both cases (AP and SS heatmaps), the network is trained from scratch for 400 epochs using Adam \cite{adam} as the optimizer and with a fixed learning rate of $10^{-4}$. Particularly, for the AP version, it is unfeasible to compute the target heatmaps using all the pixels in the images. Thus, we employ 250 keypoints, randomly sampled from the image, in the same way as in the descriptor training.

\subsection{Testing Setup and Evaluation Metrics}

For the evaluation step, we use four separate datasets:

    \textbf{FIRE} \cite{fire} is the standard in color fundus registration as it is the only dataset with registration ground-truth.  FIRE comprises 134 registration pairs across three categories: S (71 pairs) with high overlap, P (49 pairs) with low overlap, and A (14 pairs) with high overlap but pathology progression within the registration pair, making registration complex but very clinically relevant.
    
    \textbf{LongDRS} \cite{ldrs} is a diabetic retinopathy screening dataset with 1,120 images from 70 patients. Each patient has 16 images: four per eye taken across two visits, one year apart.   This dataset has no registration ground truth. The overlapping between the images is very varied, from extremely low to very high. From the total 1120 images a set of 3141 image pairs for registration can be created \cite{rivas4}.

    \textbf{DeepDRiD} \cite{deepdrid}, is a dataset originally designed for diabetic retinopathy assessment. It is comprised of 2,000 fundus images from 500 patients, with macula-centered and optic-disc-centered views for both eyes. Due to variations in the actual image centers, overlap between registration pairs varies significantly. Using these two separate viewpoints we can construct the registration pairs. From the 1,000 potential registration pairs, 10 were excluded as unsuitable for registration \cite{rivas4}.

\begin{figure}
    \centering
    \includegraphics[width=0.6\linewidth]{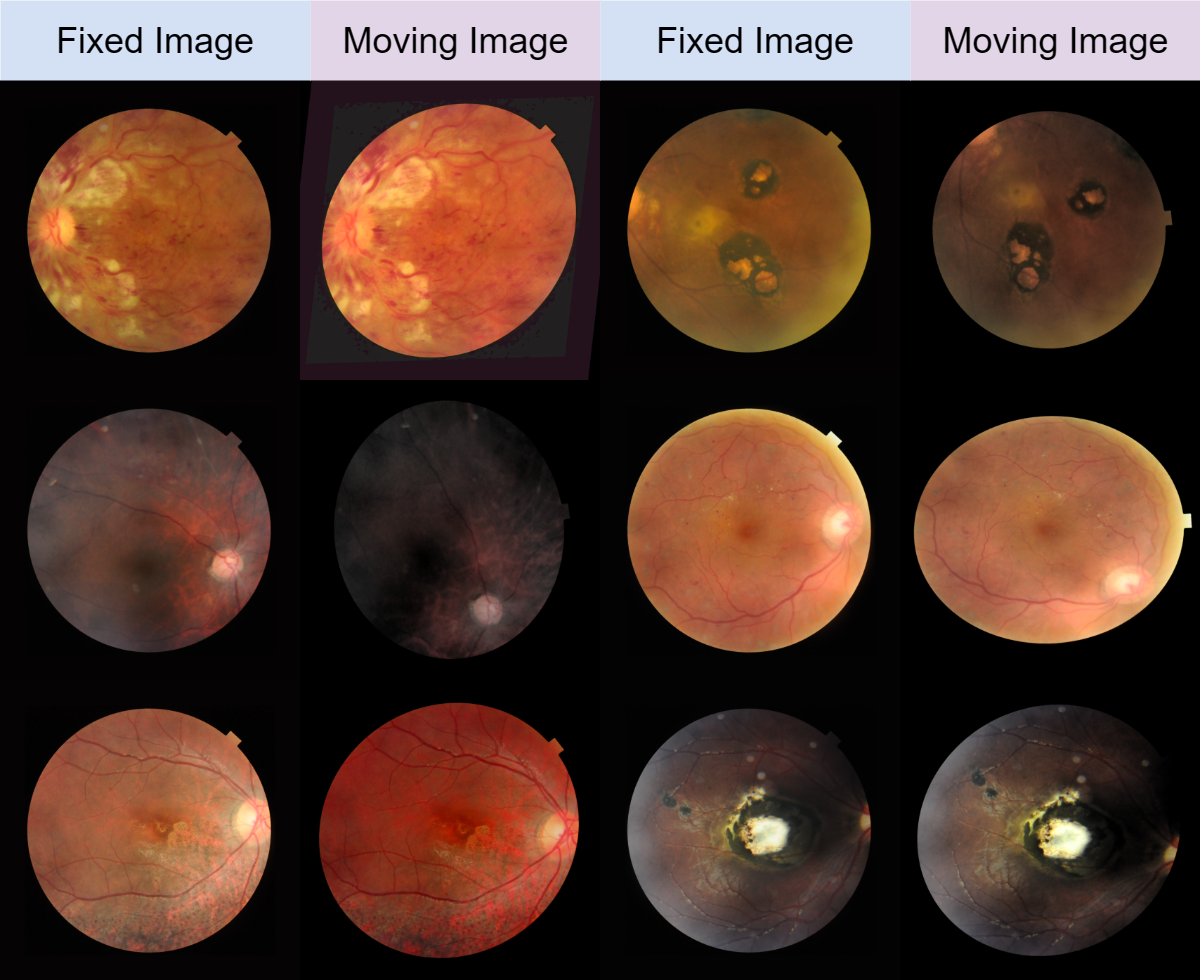}   
    \caption{Examples of fixed and moving pairs from RFMID generated via geometric and/or color augmentations.}
    \label{fig:ex_RFMID}
\end{figure}

    \textbf{RFMID} \cite{rfmid} is a dataset from a retinal screening, comprising both healthy and pathological images, including 45 different types of pathologies. The dataset has 3200 images captured by a TOPCON 3D OCT-2000, a Kowa VX-10$\alpha$ and a TOPCON TRC-NW300. Since the Kowa has a FOV of 50º instead of 45$^\circ$, like the rest of the images in this dataset and in the other ones used in this work, they were removed, for the sake of simplicity. However, it should be noted that any registration method trained in images with a FOV of 45º could be easily adapted to any other FOV \cite{rivas2}. After removing the images from the Kowa device and any other with indeterminate origin, there are a total of 2704 images left. As RFMID has no repeated visits, it does not have any registration pairs. Thus, to create pairs, we use synthetic transformations. These allow to granularly test different properties of the detector and descriptor evaluating the impact of the different transformations. In this regard, we create three separate transformations: only color changes, only geometric, both combined. 
    The color augmentations are the same ones used in the training process \cite{hsv}. The spatial augmentations are random affine transformations with rotations of up to $\pm 45^{\circ}$, scaling between $(0.9-1.1) \times imageSize$ and shearing of $\pm 10^{\circ}$. For the combined color-spatial transformation, parameters from both prior random transformations are kept so that the results are directly comparable. Figure \ref{fig:ex_RFMID} shows representative examples of fixed and moving images for this test dataset, including both color and geometric augmentations.

In terms of evaluation, as every dataset has different characteristics, we propose to use different metrics. Firstly, for FIRE and RFMID we use Registration Score, a metric based on distance between control points among the fixed and moving image, as defined in \cite{fire}. 
Following the standard benchmark, we use 25 pixels as the maximum error threshold for both FIRE and RFMID.
FIRE already has a set of ground truth control points that enable the calculation of the Registration Score, however, RFMID does not. Since the transformations in RFMID are synthetic and known, we can automatically create a set of keypoints to evaluate the registration. FIRE has a sparse set of 20 control points, which we improve in RFMID by creating 5000 control points within the retina's RoI, obtaining a more representative and accurate evaluation.

Since LongDRS and DeepDRiD lack ground truth or known transformations, Registration Score cannot be used as an evaluation metric. Thus, we employ the surrogate metrics used in \cite{rivas4}. On the one hand, DICE and Intersection Over Union (IoU) measure the registration following a segmentation of the blood vessels in both images. These are densely scattered across the whole retinal surface, offering a reliable way of measuring registration performance. Similarly, \cite{rivas4} proposed Intersection over Minimum (IoM) to address limitations of IoU and DICE in evaluation. Moreover, in \cite{rivas4}, they also use SSIM and just the Structure Metric (SM) of SSIM, albeit with a slight modification to ensure a reliable evaluation of registered images \cite{rivas4}. Finally, following \cite{rivas4}, we also use LPIPS for evaluation, employing the VGG version, which outperforms the AlexNet version \cite{rivas4}. Importantly, due to the high number and complexity of image pairs, some approaches may fail to register all pairs. To address this, we normalize the metrics based on the number of registrations, enabling direct comparison regardless of the registration count \cite{rivas4}. However, the registration count is also a key metric in these datasets.

Finally, in addition to evaluating the full registration pipeline, we propose to evaluate the detector's performance using the RFMID dataset, which has known transformations. This allows us to quantify detector accuracy by measuring the average Euclidean distance between matched keypoints.

\section{Results and Discussion}
\subsection{Unsupervised Descriptor}

First, we compare the performance of the proposed unsupervised descriptor network, UnConKeD, and the supervised counterpart of the reference work, ConKeD++ \cite{rivas4}. To do so, we use the supervised keypoint detector from  ConKeD++. This detector identifies blood vessel crossovers and bifurcations as keypoints which are then employed in the descriptor training in ConKeD++. Therefore, this setting represents the most favorable scenario for the supervised description method, as we are using the specific points used to train it. 

The results for this experiment are shown in Table \ref{tab:desc}. In this table we can see that, despite the unsupervised training, UnConKeD produced descriptors able to match or even outperform the supervised approach. The unsupervised descriptor approach trains with more diverse samples (keypoints), enabling it to accurately describe a broader set of keypoints. In contrast, the supervised approach is tailor-made for blood vessel crossovers and bifurcations. Consequently, in the supervised approach, if the detector mistakenly selects a point that is neither a crossover nor a bifurcation, the descriptor may produce inaccurate descriptors leading to incorrect matching, since it only received training feedback for crossovers and bifurcations. However, given the scarcity of blood vessels, especially crossovers and bifurcations, on the retinal surface, consistently sampling these points for training in the unsupervised approach is highly unlikely, whereas, in directed sampling these are the only points used. Despite this, and the use of crossovers and bifurcations as keypoints for this experiment, the unsupervised approach obtains similar or better metrics across the three test datasets.

In terms of specific numeric results, in FIRE, the unsupervised method improves the overall performance despite decreasing it slightly in Category A. This is offset by improved performance in Categories P and S. In LongDRS \cite{ldrs}, the unsupervised descriptors are able to pair less images. However, it improves most metrics (only LPIPS worsens), despite being heavily penalized by the lower number of registered pairs, indicating a notable improvement in registration accuracy. In contrast, in DeepDRiD it registers more images and does so also improving the metrics, signaling a clear, all-around improvement in registration quality. Therefore, we can conclude that the unsupervised approach is superior, as it achieves performance that is comparable to or better than the supervised approach. Notably, our proposal avoids the performance loss typically associated with unsupervised approaches, achieving this even in the most favorable setting for the supervised approach. Unsupervised methods generally offer a trade-off, sacrificing performance for improved adaptability compared with supervised methods. This is especially important given the scarcity of labeled datasets in the medical domain, and even more so for datasets that include ground truth information relevant to registration, such as keypoints.

\begin{table}[]
\centering

\begin{tabularx}{0.85\textwidth}{lYYYYYY}
\toprule
\mbox{\textbf{FIRE}}                              & FIRE           & A              & P              & S              & Avg.           & \mbox{W. Avg.}                        \\ \midrule
\mbox{ConKeD++ \cite{rivas4}} & 0.764          & \textbf{0.766} & 0.503          & 0.945          & 0.738          & 0.765                         \\
\mbox{UnConKeD}                           & \textbf{0.769} & 0.757          & \textbf{0.513} &  \textbf{0.948} & \textbf{0.739} & \textbf{0.769}                 \\ \midrule
\end{tabularx}

\begin{tabularx}{0.85\textwidth}{lYYYYYYY}

\textbf{LongDRS}                           & Pairs       & IoU            & DICE           & IoM            & SM             & SSIM           & LPIPS          \\ \midrule
\mbox{ConKeD++ \cite{rivas4}} & \textbf{2821}  & 0.518          & 0.323          & \textbf{0.673} & 0.627          & 0.620          & \textbf{0.193} \\
\mbox{UnConKeD}                              & 2811           & \textbf{0.520} & \textbf{0.324} & \textbf{0.673} & \textbf{0.677} & \textbf{0.671} & 0.196          \\ \midrule
\textbf{DeepDRiD}                         & Pairs       & IoU            & DICE           & IoM            & SM             & SSIM           & LPIPS          \\ \midrule
\mbox{ConKeD++ \cite{rivas4}} & 905            & 0.481          & 0.311          & 0.658          & 0.603          & 0.588          & 0.253          \\
\mbox{UnConKeD}                               & \textbf{914}   & \textbf{0.485} & \textbf{0.313} & \textbf{0.665} & \textbf{0.647} & \textbf{0.632} & \textbf{0.245} \\ \bottomrule
\end{tabularx}%
\vspace{1em}
\caption{Comparison of the results between supervised descriptor training (ConKeD++) and unsupervised descriptor training (UnConKeD). For LongDRS and DeepDRiD the metrics are normalized given the number of registered pairs.}
\label{tab:desc}

\end{table}

\subsection{Unsupervised Keypoint Detection}

\subsubsection{Comparison of candidate heatmaps} \label{sec:comp-heatmaps}

In this experiment, we study the performance of the three explored heatmap construction approaches: AP, SS and the combination of both. Results are shown in Table \ref{tab:maps5h}. Table \ref{tab:maps5h}-a shows the performance using up to 200 keypoints while Table \ref{tab:maps5h}-b, up to 500 keypoints. In these Tables we can see that both heatmaps offer accurate performance. Similarly, the combination of both heatmaps also performs satisfactorily, even improving the results of each one of them separately in some cases. This is the case when employing 200 keypoints, as the combined approach performs the best across the majority of datasets and metrics, although it is closely followed by the Self-Similarity map. Increasing the number of keypoints to 500 benefits SS to a greater extent than AP, and it obtains the best overall performance. The combination of both heatmaps appears to be hindered by the lower performance of the AP heatmap.

Based on the results obtained from the four separate datasets and across the two different numbers of keypoints, we select the Self-Similarity heatmap as the most effective approach for estimating descriptor performance. While the addition of the AP loss has the potential to enhance performance (i.e., at 200 keypoints), the results in the case of 500 keypoints show that this is not guaranteed. Thus, the added computational cost of creating both heatmaps is not justified.

\begin{figure}[tb]
    \centering
    \includegraphics[width=0.65\textwidth]{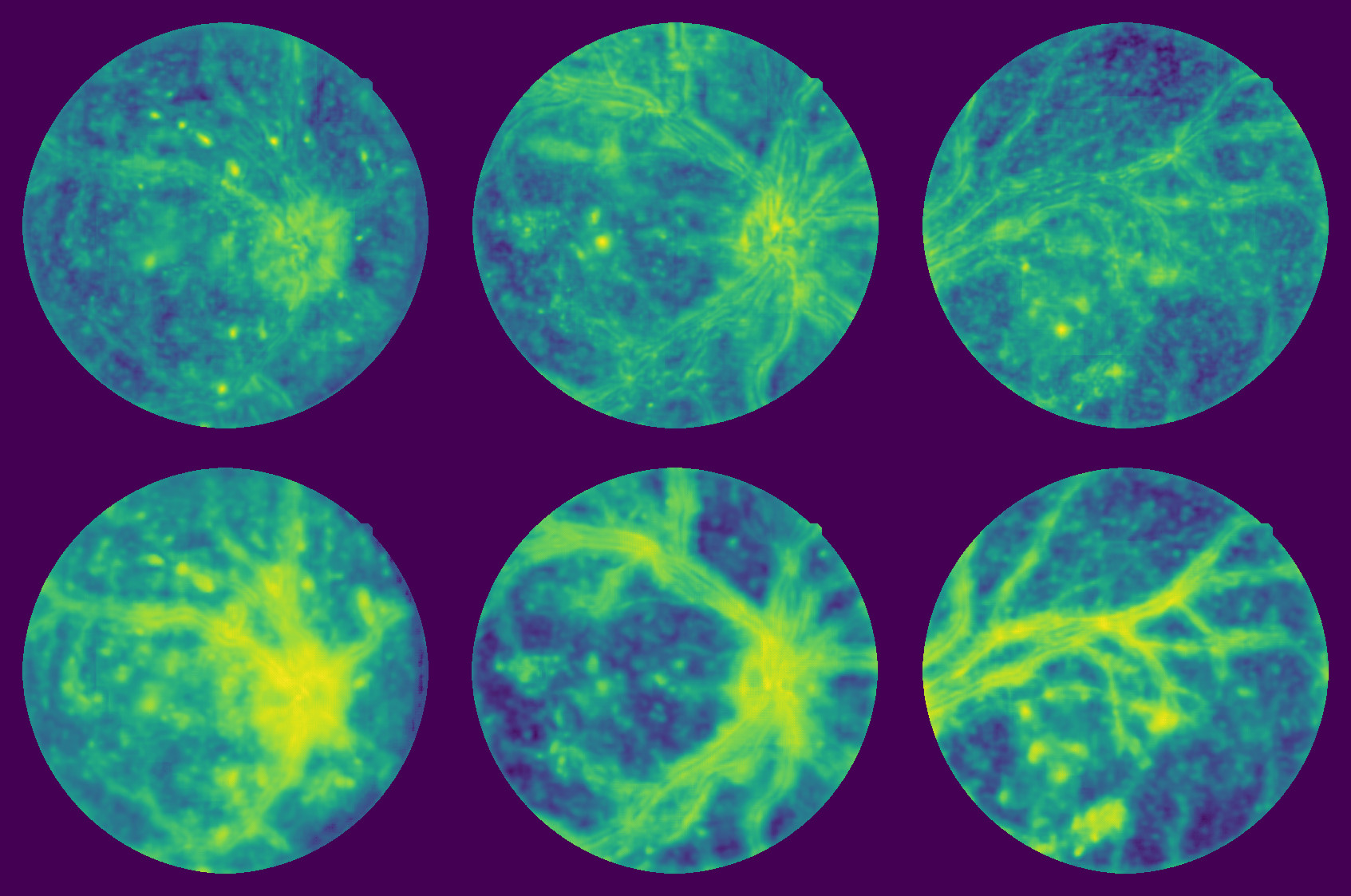}
    \caption{Heatmaps from the hold-out test dataset FIRE. Top: SS. Bottom: AP (inverted for comparison). Higher values (i.e., yellow) indicate better areas.   
    }
    \label{fig:hm-ex}
\end{figure}

\begin{figure*}
    \centering
    \includegraphics[width=\linewidth]{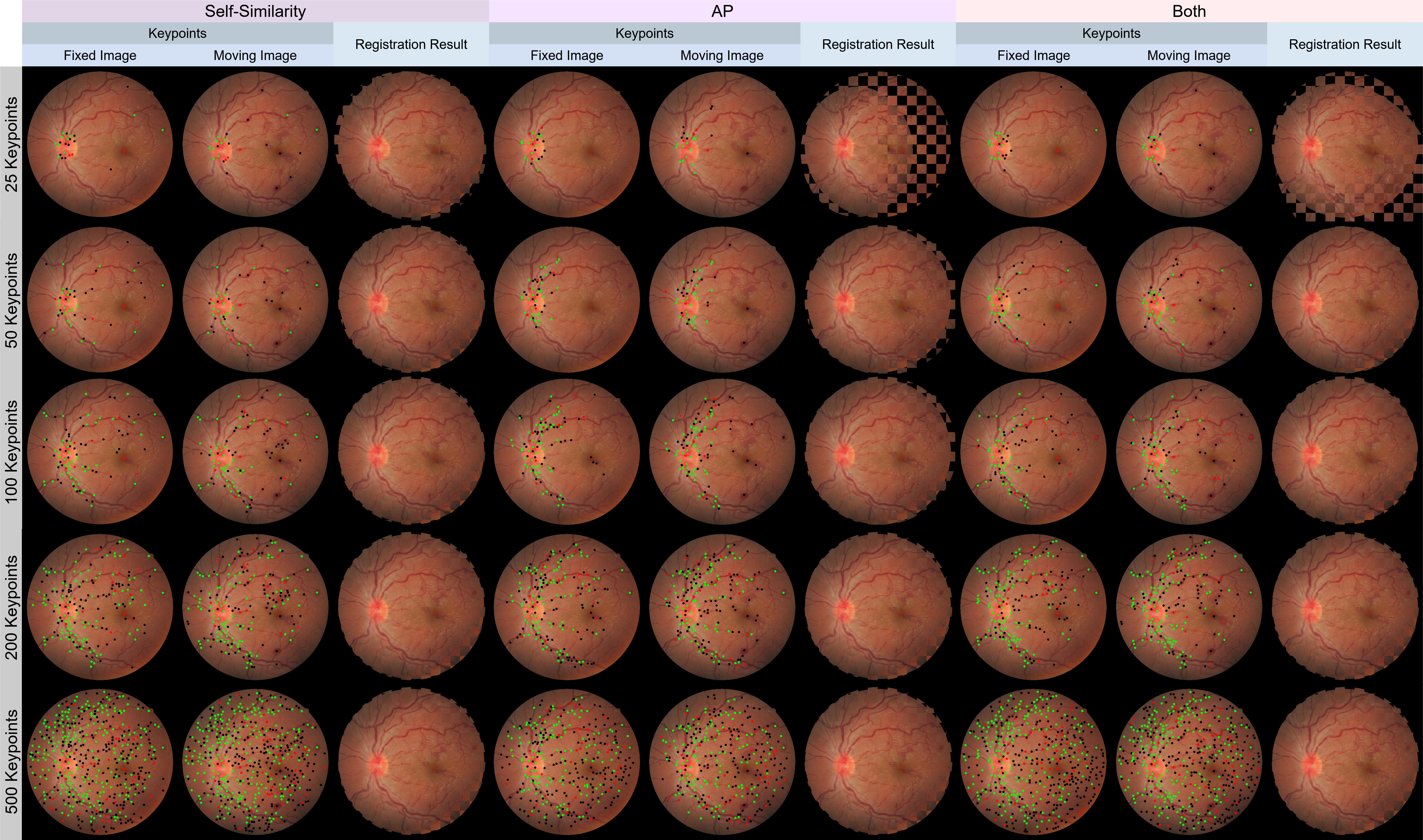}
    \caption{Representative registration example from the FIRE dataset (Category A) hold-out test set, illustrating performance across varying keypoint counts and heatmap sources. Green keypoints represent successful inlier matches, red keypoints denote matched outliers rejected by RANSAC, and black keypoints indicate unmatched features.}
    \label{fig:kps1}
\end{figure*}

Relevant examples of the heatmaps created using SS and AP are shown in Figure \ref{fig:hm-ex}. In these examples we can see that, while both types of heatmaps capture a similar structure, closely following key representative patterns for registration, such as blood vessels or lesions, the Self-Similarity map has many more details. The extra details allow for finer keypoint localization, which explains the improved performance of the SS heatmap, especially at higher numbers of keypoints. Additionally, Figure \ref{fig:kps1} shows a representative example from the FIRE hold-out test set, depicting the keypoints extracted from the different heatmap-constructing approaches, varying the number of detected keypoints in the same way as the experiments presented in this section, \ref{sec:comp-heatmaps}, and the following one, \ref{sec:m-eff}. It should be noted that the number of keypoints is the average over the FIRE dataset, so some variation may occur in individual images. The different properties of the AP and SS heatmaps are clear in the presented images, although they vary depending on the number of keypoints.

Firstly, at the lower levels of keypoints, we can see how SS is able to consistently detect non arterio-venous related keypoints, such as the macula. This is relevant, as typical crossover and bifurcation detectors are unable to detect any keypoints in this region due to its avascular nature. This behavior is only exhibited by AP at higher numbers of keypoints. At the 200-keypoint level, where the combined heatmap outperformed both AP and SS, we can see that the AP closely follows the arteriovenous tree whereas SS includes more non-vascular points. The combined heatmap provides a compromise between both as it includes both many vascular points and non-vascular points. This compromise is key at this keypoint level since including points in peripheral or non-represented zones in the retina can help regularize the registration in those parts. In this case, the compromise provided by the combined heatmap is beneficial as it includes both types of keypoints without over-representing any group, which would not improve registration results as much. 

Conversely, at the 500 keypoint-level, as the number of relevant vascular keypoints saturates in both the AP and SS heatmaps, the benefits of the combined approach diminish. In this scenario, the higher ability of the SS heatmap to find high-quality non-vascular points is more significant, resulting in better overall performance.  This is a direct consequence of the heatmaps shown in Figure \ref{fig:hm-ex}, where the AP heatmaps exhibit significantly less details, having a wide number of high-intensity points in vascular zones, leading to a preference for generating vascular-related keypoints.

\begin{table*}[]

\begin{tabularx}{1\textwidth}{lYYYYYYYYYYYY}

\cmidrule(r){2-7}\cmidrule(r){8-13}

\textbf{} & \multicolumn{6}{c}{a) \textbf{Results for  200 keypoints}} & \multicolumn{6}{c}{b) \textbf{Results for 500 keypoints}} \\

\cmidrule(r){1-1}\cmidrule(r){2-7}\cmidrule(r){8-13}

\textbf{FIRE} & FIRE & A & P & S & Avg. & \mbox{W. Avg.} & FIRE & A & P & S & Avg. & \mbox{W. Avg.} \\ \cmidrule(r){1-1}\cmidrule(r){2-7}\cmidrule(r){8-13}

AP Loss & 0.628 & 0.546 & 0.304 & 0.868 & 0.573 & 0.628 & 0.685 & 0.674 & 0.385 & 0.895 & 0.651 & 0.685 \\
Self-Similarity & 0.636 & 0.531 & 0.311 & 0.881 & 0.574 & 0.636 & 0.719 & \textbf{0.731} & 0.429 & \textbf{0.917} & 0.692 & 0.719 \\
Combined & \textbf{0.659} & \textbf{0.589} & \textbf{0.353} & \textbf{0.885} & \textbf{0.609} & \textbf{0.660} & \textbf{0.730} & 0.729 & \textbf{0.461} & 0.915 & \textbf{0.702} & \textbf{0.730} \\ \cmidrule(r){1-1}\cmidrule(r){2-7}\cmidrule(r){8-13}

\end{tabularx}

\par\vskip0pt
\begin{tabularx}{1\textwidth}{lYYYYYYYY} 

\textbf{RFMID} & All & Full & Color & Geo. & All & Full & Color & Geo. \\ 
\cmidrule(r){1-1}\cmidrule(r){2-5}\cmidrule(r){6-9}

AP Loss & 0.944 & 0.925 & 0.939 & 0.969 & 0.952 & 0.932 & 0.946 & 0.977 \\
Self-Similarity & 0.961 & 0.944 & 0.953 & \textbf{0.985} & \textbf{0.976} & \textbf{0.963} & \textbf{0.969} & \textbf{0.995} \\
Combined & \textbf{0.963} & \textbf{0.947} & \textbf{0.956} & \textbf{0.985} & 0.973 & 0.959 & 0.968 & 0.994
\\ \cmidrule(r){1-1}\cmidrule(r){2-5}\cmidrule(r){6-9}

\end{tabularx}%

\begin{tabularx}{1\textwidth}{lYYYYYYYYYYYYYY}
\textbf{LongDRS} & Pairs & IoU & DICE & IoM & SM & SSIM & LPIPS & Pairs & IoU & DICE & IoM & SM & SSIM & LPIPS \\ \cmidrule(r){1-1}\cmidrule(r){2-8}\cmidrule(r){9-15}

AP Loss & 2292 & 0.359 & 0.231 & 0.478 & 0.625 & 0.630 & 0.369 & 2453 & 0.380 & 0.245 & 0.507 & 0.633 & 0.638 & 0.325 \\
Self-Similarity & \textbf{2546} & 0.397 & \textbf{0.256} & \textbf{0.531} & \textbf{0.637} & 0.642 & \textbf{0.295} & \textbf{2868} & \textbf{0.475} & \textbf{0.301} & \textbf{0.627} & \textbf{0.661} & \textbf{0.668} & \textbf{0.196} \\
Combined & 2533 & \textbf{0.398} & \textbf{0.256} & \textbf{0.531} & \textbf{0.637} & \textbf{0.643} & 0.299 & 2810 & 0.454 & 0.289 & 0.6 & 0.655 & 0.661 & 0.216 \\ \cmidrule(r){1-1}\cmidrule(r){2-8}\cmidrule(r){9-15}
\end{tabularx}%

\begin{tabularx}{1\textwidth}{lYYYYYYYYYYYYYY}
\textbf{DeepDRiD} & Pairs & IoU & DICE & IoM & SM & SSIM & LPIPS & Pairs & IoU & DICE & IoM & SM & SSIM & LPIPS \\ \cmidrule(r){1-1}\cmidrule(r){2-8}\cmidrule(r){9-15}
AP Loss & 927 & 0.335 & 0.236 & 0.503 & 0.598 & 0.604 & 0.244 & 932 & 0.340 & 0.239 & 0.508 & 0.595 & 0.606 & 0.241 \\
Self-Similarity & 944 & 0.335 & 0.237 & 0.507 & 0.594 & 0.604 & 0.230 & \textbf{963} & \textbf{0.358} & \textbf{0.250} & \textbf{0.534} & \textbf{0.599} & \textbf{0.610} & 0.215 \\
Combined & \textbf{950} & \textbf{0.345} & \textbf{0.243} & \textbf{0.519} & \textbf{0.596} & \textbf{0.606} & \textbf{0.225} & \textbf{963} & 0.357 & 0.249 & 0.532 & \textbf{0.599} & \textbf{0.610} & \textbf{0.214}
\\ \cmidrule(r){1-1}\cmidrule(r){2-8}\cmidrule(r){9-15}
\end{tabularx}%
\caption{Comparison of the results between the three different heatmap approaches used for keypoint detection, across the four datasets using 200 (a) and 500 (b) keypoints. For FIRE and RFMID results are measured in Registration Score. For LongDRS and DeepDRiD the metrics are normalized given the number of registered pairs. The best results for each combination of number of keypoints and dataset are highlighted in bold.}
\label{tab:maps5h}

\end{table*}


\subsubsection{Matching Efficiency and Number of Keypoints} \label{sec:m-eff}

Once the best heatmap is selected (i.e., SS), we can evaluate its performance across a wider range of keypoint quantities. Using the SS map reduces the number of possible combinations that would otherwise need to be tested if all three original heatmap alternatives were considered, making the process more efficient and focused. In particular, the goal of this experiment is to evaluate how accurate and distinctive our keypoint detector is and how well it performs coupled with the descriptor block.

In this section, we also compare our detector approach to other detectors in the state of the art that can be decoupled from their descriptors. We include a comparison to the supervised detector of blood vessel crossovers and bifurcations, used in \cite{rivas3,rivas4}. This detector is able to detect domain-specific keypoints which are, by definition, very precise and highly accurate. Furthermore, we compare our detector against the detection approach presented in D2-Net \cite{d2}. This method identifies keypoints through a series of calculations applied to the computed descriptor block. It works by first performing non-local-maximum suppression on a feature map, followed by an additional non-maximum suppression across each descriptor. As a result, it functions as an unsupervised keypoint detector that relies exclusively on the computed descriptor block. However, unlike our proposed approach, which leverages a dedicated network, this method is based on heuristic computations to locate keypoints.

\begin{figure*}[htb]
    \centering
    \includegraphics[width=\textwidth]{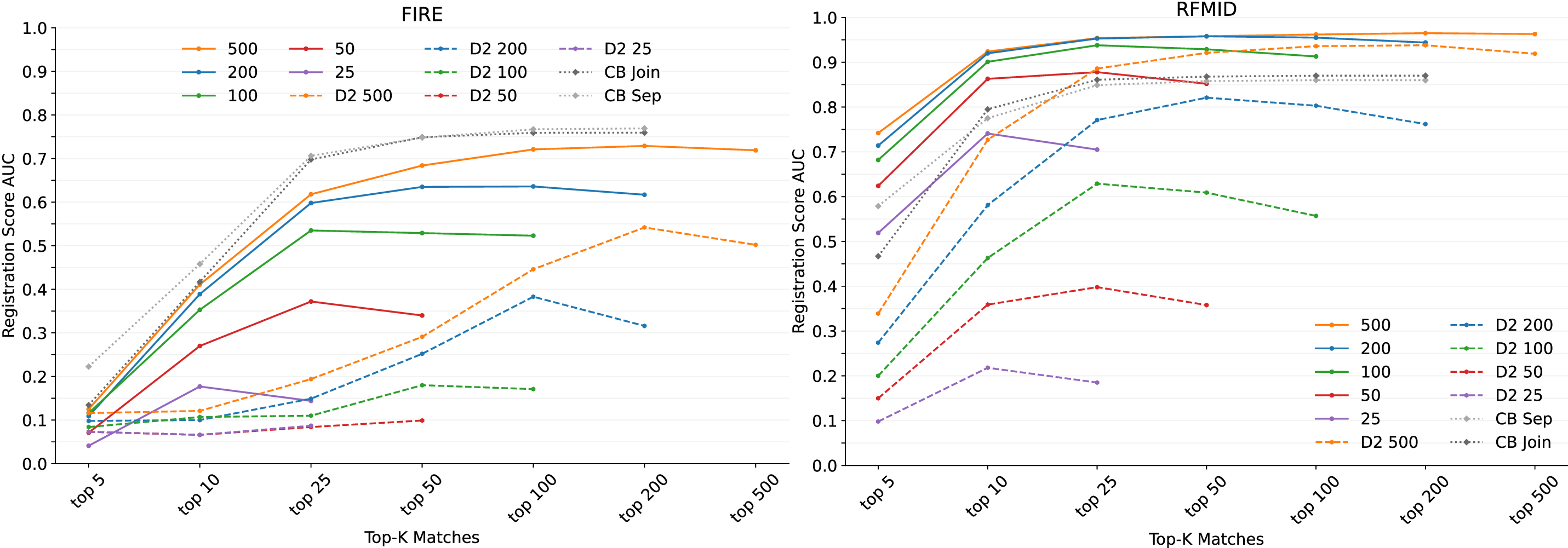}
    \caption{Results for our approach (labeled by the number of keypoints used), D2 (labeled D2 + number of keypoints) and the supervised crossovers and bifurcations (CB Join or CB Sep) in FIRE and RFMID; measured in Registration Score AUC.}
    \label{fig:res-fr}
\end{figure*}

\begin{figure*}[htb]
    \centering
    \includegraphics[width=\textwidth]{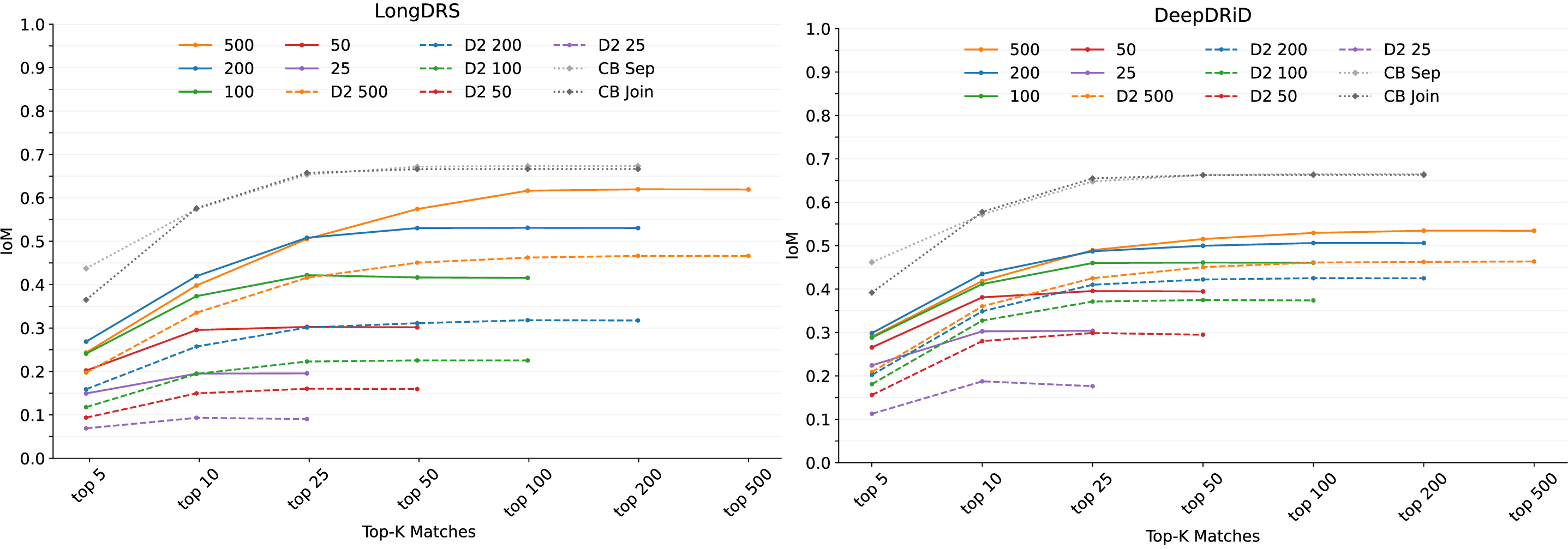}
    \caption{Results for our approach (labeled by the number of keypoints used), D2 (labeled D2 + number of keypoints) and the supervised crossover and bifurcation points (labeled CB Join or CB Sep) in LongDRS and DeepDRiD, measured in IoM.}
    \label{fig:res-ld}
\end{figure*}

The results for this experiment are shown in Figures \ref{fig:res-fr} and \ref{fig:res-ld}. Importantly, these Figures measure two factors. Each line shows the number of detections (i.e., keypoints chosen by the detector network or method). On the other hand, the X axis shows the number of matches chosen from each set of detections (line). This number of matches represents the closest descriptors, determined based on their respective pairwise distances. Logically, the number of selected matches cannot exceed the number of originally detected keypoints. This constraint is reflected in the figures, where the lines corresponding to lower numbers of keypoints end earlier along the x-axis compared to those representing higher numbers of keypoints.

Firstly, our approach outperforms D2 in every dataset and given any number of detected keypoints. This is especially notable in FIRE and LongDRS where the gap between both unsupervised approaches is the biggest. On the other hand, in DeepDRiD and RFMID, the performance gap is notably smaller compared to the other two datasets. RFMID shows a smaller gap at higher keypoint counts (500), while DeepDRiD shows less gap at lower keypoint numbers. The supervised keypoints (crossovers and bifurcations) naturally outperform both unsupervised approaches, as expected, given the dependence on a manually annotated ground truth. 

In terms of the supervised keypoints, there is a clear difference between CB Sep and CB Join. On the one hand, CB Join disregards the class of these keypoints, mixing crossovers and bifurcations, allowing matching between these two types of keypoints. This results in generally lower performance compared to CB Sep, which only matches crossovers to crossovers and bifurcations to bifurcations. This distinction is relevant because differentiating the type of keypoint improves registration but also increases the complexity of the overall method. The difference between CB Sep and CB Join, which is especially notable in the lowest number of matches selected, highlights the need for reliable matching. Separating the keypoints by class improves performance as it allows for a more precise distinction than just using descriptors. However, this performance disparity diminishes rapidly as more keypoints are selected, disappearing at 25 points in all cases. Interestingly, while in FIRE the separated approach performs better, in RFMID the joined approach produces better results. This is likely due to a keypoint that was misidentified as the other class by the detector but was still accurately matched using descriptors. Overall, these findings show that, given an adequate detection approach, an accurate descriptor block (i.e., our unsupervised descriptor) can be leveraged to obtain a precise registration, even at very low numbers of keypoints.

In the case of RFMID, our method is able to consistently outperform either of the supervised keypoint detection approaches, even with very few detected keypoints. This is likely due to the specifics of registration on RFMID which has synthetic transformations. These synthetic transformations, in principle, simplify registration for all detection approaches, as every keypoint in one image will also be present in the other.  This in turn, benefits our approach as it can choose the most relevant points in the image, which are guaranteed to be present in both components of the registration pair and will, most likely, be the exact same despite the heavy augmentations. Therefore, even with extremely few keypoints, our method manages to produce satisfactory and accurate registrations.

From the results seen in Figures \ref{fig:res-fr} and \ref{fig:res-ld} we can conclude that our approach offers better performance than current unsupervised detector approaches, represented in the form of D2. Furthermore, our method also offers performance comparatively similar to state-of-the-art domain-specific supervised keypoints (blood vessel crossovers and bifurcations) albeit with a generally higher number of keypoints. This is especially notable in FIRE and LongDRS where our approach is closer to the supervised version. On DeepDRiD the gap between our unsupervised method and the domain-specific keypoints widens. On the other hand, in RFMID, the unsupervised approach performs better than the supervised methods. In any dataset other than RFMID, our approach yielded results closer to those of the supervised method (be it joined or separated by class) when employing 500 keypoints. This number of keypoints is significantly higher than those typically detected by the supervised method, which usually identifies over 100 keypoints, labeled in the figures as CB Sep or CB Join. However, supervised methods are limited by the number of possible keypoint detections, as there is a finite number of detectable crossovers and bifurcations. 
In contrast, our heatmap-based approach can sample an arbitrary number of keypoints, enabling a trade-off between efficiency and registration accuracy.

Analyzing the performance of our method in isolation, we can see that, in most cases, we could select as low as 1/4 of the matches, using the matching distance between descriptors as criteria, and still obtain around the same performance as with all of them. This highlights that, given accurate detection and matching, most matches are redundant in order to create a transformation. However, due to the nature of registration, more keypoints are still useful, especially when they are spread over the image as they regularize the transformation in those locations. This helps in creating transformations that map the deformation realistically and to a fine degree, slightly improving the numerical results.

\begin{figure*}[htb]
    \centering
    \includegraphics[width=\textwidth]{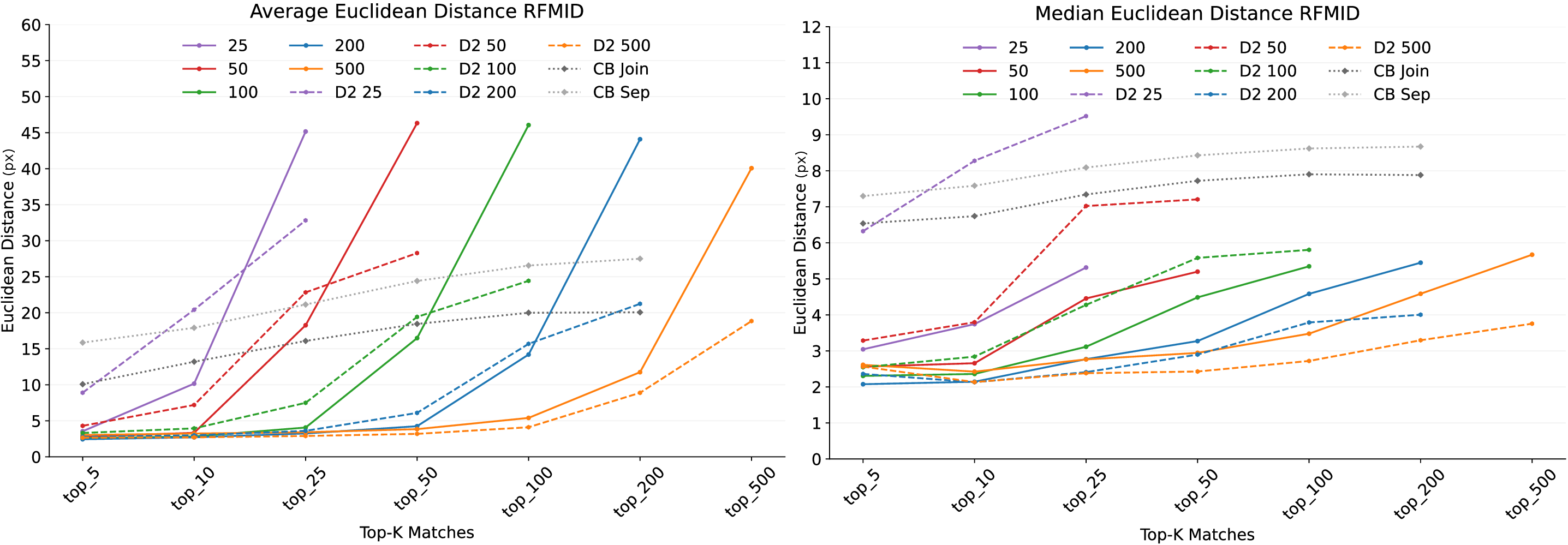}
    \caption{Results of the keypoint detection distance experiment, comparing our approach (labeled by the number of keypoints used), D2 (labeled D2 + number of keypoints) and the supervised crossover and bifurcation points (labeled CB Join or CB Sep) using as metric the Euclidean distance in the image space, measured in pixels.}
    \label{fig:dists}
\end{figure*}

Finally, we evaluate the performance of our detector using the Euclidean distance between matched keypoints in the image space (i.e., pixels). Specifically, we compute the distance between matched keypoint pairs after re-projecting the moving image onto the fixed image space. Ideally, a perfect match would result in distances approaching zero, though minor deviations may arise due to interpolation artifacts. Larger distances indicate greater localization errors, with substantial discrepancies suggesting incorrect matches. These results are shown in Figure \ref{fig:dists}. In this Figure we can see that, for every method, using all the matches increases the keypoint distance. In general, selecting the matches based on descriptor distance effectively discriminates the better localized keypoints.

The supervised methods have a steady growth of the average and median distances. However, this tendency is much more pronounced in the unsupervised methods. Notably, until approximately 50\% of the matches are selected the unsupervised methods have lower distance than the supervised ones. Interestingly, we can also see that our method shows more error when detecting more keypoints (i.e., 500) than the D2 approach. This, as hinted by the fact that D2's registration performance is notably worse than our approach, is because D2 detects keypoints all within similar locations (like the optic disc). This causes mismatches to be punished less as they are physically closer even if incorrectly matched. However, the tightly grouped points significantly degrade registration performance as there are not enough points in the more peripheral regions of the retina to ensure accurate registration. Additionally, since D2 completely fails to register some images, its results are unfairly boosted. 

Importantly, the notable difference between the average and the median for our method suggests the presence of some erroneous localizations or matches that harshly penalize it in this test. Analyzing the median distance provides a more robust evaluation of the methods, as it is resistant to outliers on either end, both exceptionally close keypoints and significantly mismatched ones. In terms of the median distance, unsupervised methods generally outperform supervised ones, highlighting the difficulty of precisely identifying crossover or bifurcation points.  This, in part, stems from the inherent limitations of human-generated labels, which may introduce ambiguities and/or biases in ground truth annotations. Both unsupervised methods trade places, with our method producing less distance when using fewer keypoints and D2 producing better results when using more keypoints. As noted previously, the spatial clustering of D2’s detections improves its performance in these tests. Overall, these results underscore the complementary nature of our different evaluations, as the different metrics help characterize the behavior and the expected performance of each method.

Notably, the supervised keypoints offer worse performance than the unsupervised ones, especially if the median is the reference metric. This is due to several factors.  Firstly, while the rough identification of domain-specific keypoints in a supervised manner may be better than the unsupervised ones, fine details may be more complex. As supervised points are labeled by humans, the exact point that is a crossover or a bifurcation may vary between images or individuals labeling the images. Thus, fluctuations, even in ground truth points, are to be expected, preventing pixel-perfect matching. Then, these biases and errors in the ground truth are transferred to the detector network itself and manifest in the distribution of its keypoints. Moreover, crossovers and bifurcations are typically scattered around the retina.  In contrast, unsupervised keypoints can capture a broader range of salient features, including additional points such as curvatures or lesions. As a result, unsupervised keypoints may be more densely clustered. Consequently, supervised points are more likely to be harshly penalized for mismatches in this evaluation due to the higher image-space distances between keypoints. Finally, both D2 and our proposed method are descriptor-based, which, expectedly, increases the likelihood that they adjust more effectively with the descriptors’ matching capabilities.

\begin{figure}[b!]
    \centering
    \includegraphics[width=0.85\linewidth]{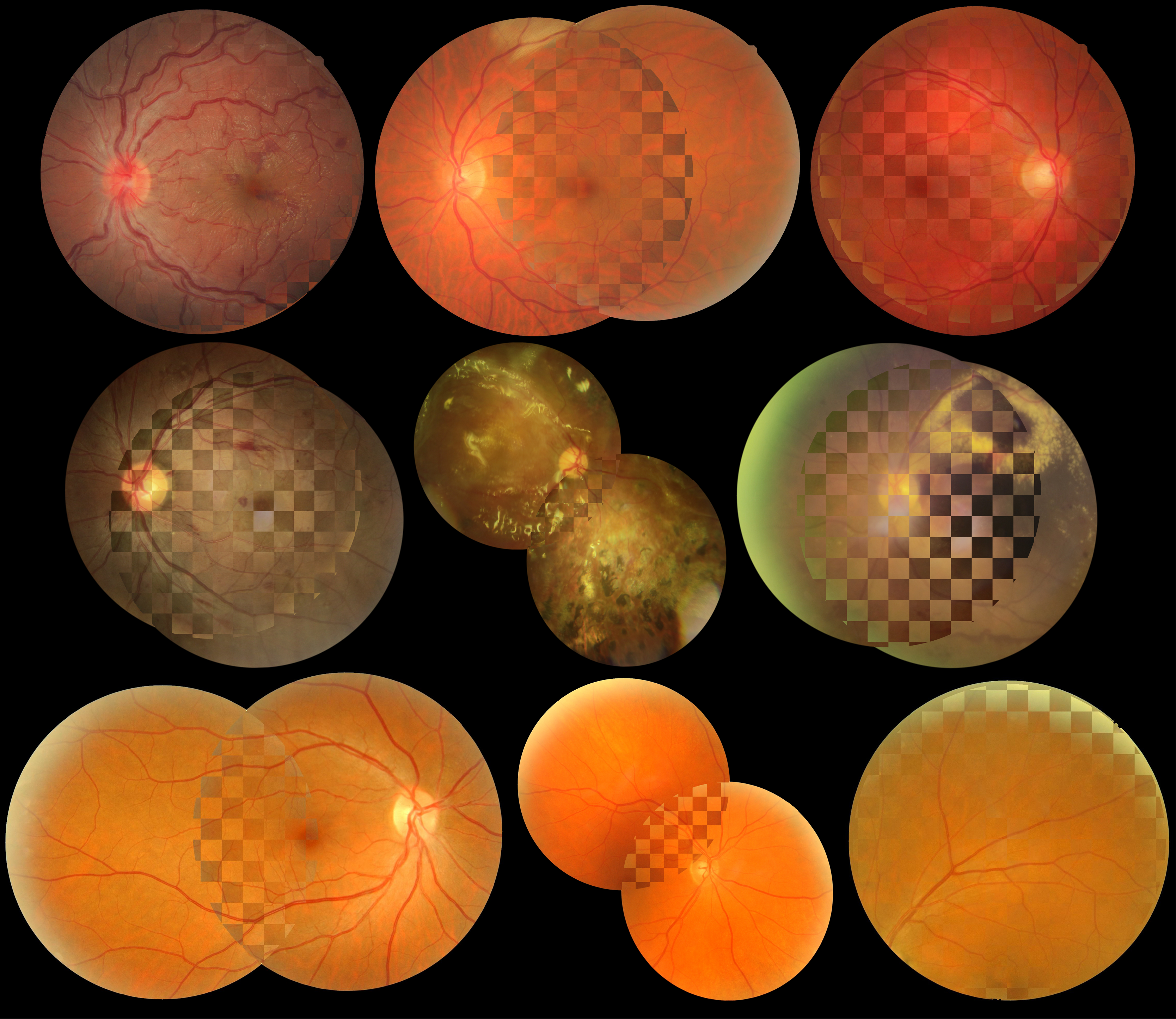}
    \caption{Representative registration examples using the proposed approach across the test datasets, illustrating a range of difficulty levels and diverse scenarios. Top row: FIRE \cite{fire} (left to right) categories A (pathological progression), P (low overlap), and S (high overlap). Middle row: DeepDRiD \cite{deepdrid}, demonstrating variability in image viewpoints. Bottom row: LDRS \cite{ldrs}, the first two images (left to right) show intra-visit registration with low overlap, while the third illustrates inter-visit registration, with a one-year interval between acquisitions, in a location different than the optic disk area.}
    \label{fig:set_ex}
\end{figure}

Finally, we include representative registration examples using the selected approach on the FIRE, LDRS, and DeepDRiD datasets in Figure \ref{fig:set_ex}. These examples illustrate the diversity of scenarios encountered across the test datasets, both in terms of geometric deformation, the presence of disease, as well as the varying locations of color fundus images across the retina. It is important to note that RFMID was excluded from this demonstration because its images are synthetically deformed. As a result, accurately registered images would appear nearly identical to the fixed images, aside from potential color variations.

\subsection{State of the Art comparison}

\begin{figure}[t]
    \centering
    \includegraphics[width=0.85\linewidth]{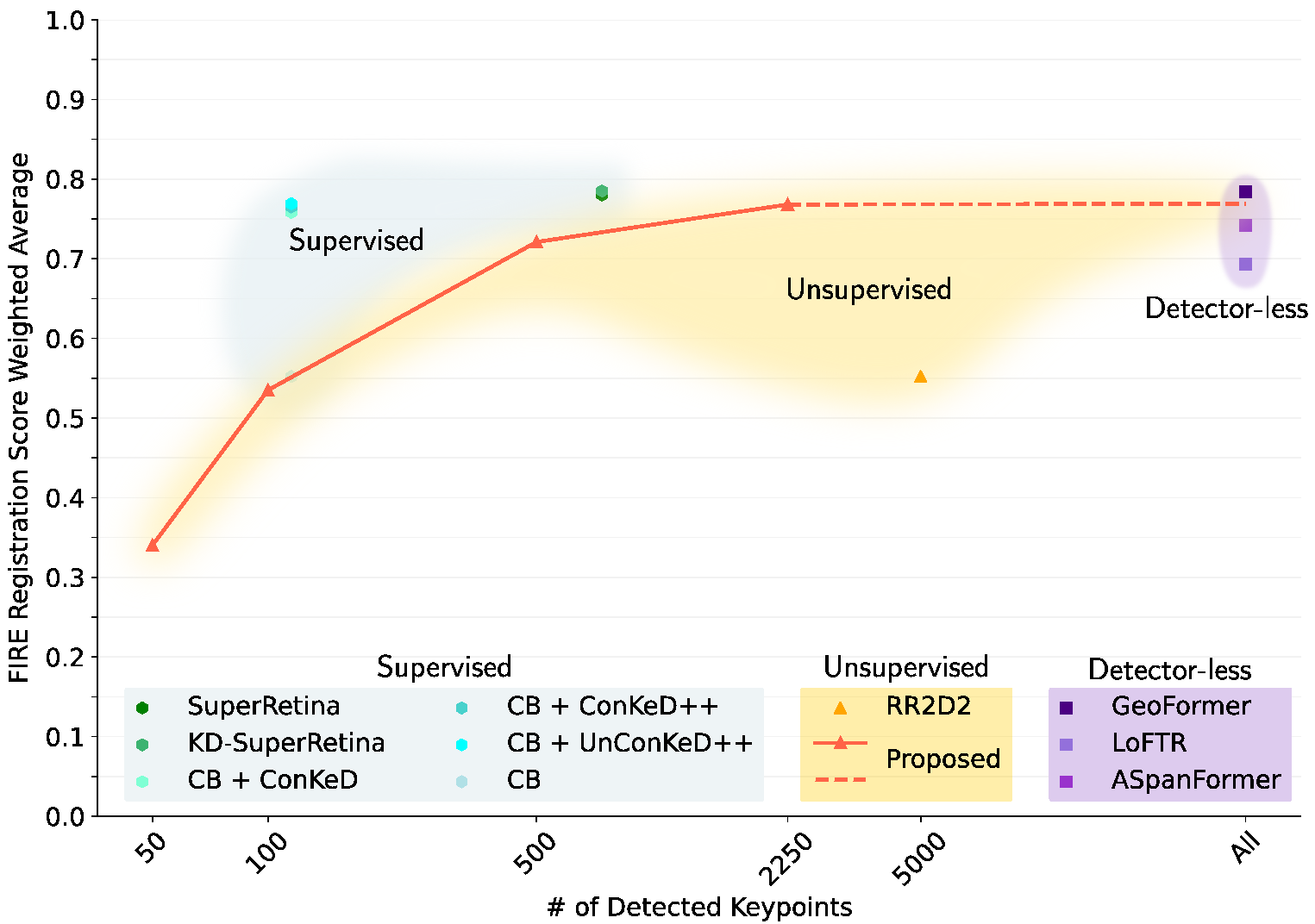}
    \caption{State-of-the-art deep learning results in FIRE, measured by the weighted average of Registration Score across the whole dataset, compared with the number of detected keypoints.}
    \label{fig:sota}
\end{figure}

We compare our method against state-of-the-art approaches using the FIRE dataset, the standard for CF registration,  in Table \ref{tab:sota}. It is important to note that this table categorizes the methods by type, facilitating a clearer comparison between different approaches to registration. Furthermore, this comparison can be intuitively visualized in Figure \ref{fig:sota} which considers the number of keypoints used by each method, a crucial factor to evaluate their relative performance and efficiency. In Table \ref{tab:sota}, we present our fully unsupervised methodology at two keypoint levels, while in Figure \ref{fig:sota}, we leverage the flexibility afforded by our heatmap-based detector, which allows us to sample an arbitrary amount of keypoints, by including a continuous curve depicting the performance of our method at multiple numbers of keypoints. Additionally, we also include in these comparisons our unsupervised descriptor paired with crossovers and bifurcations. Finally, it is important to note that some methods deviate slightly from the standard testing approach, as marked by *. In category P, a control point in one of the images is mislabeled. Although we decided to simply evaluate this image with one fewer control point, several other methods completely excluded that image from their evaluation. This results in scores that are not fully comparable, impacting both category P and the overall aggregate metrics.

Overall, our method achieves the best results among the unsupervised detector-based methods and drastically closes the gap to supervised deep learning methods. However, the classical VOTUS method \cite{votus} is still the best performing approach in the state of the art.  The uneven image distribution in FIRE causes VOTUS to rank first due to its excellent results in category P, despite not leading in categories A and S. This method has a transformation model with more degrees of freedom than any other method in the state of the art, thus, it obtains the best results in Category P which has the largest expected transformations. The drawback of larger deformation models is that they can produce unrealistic transformations in the constrained setting of CF images.  Similarly, the best approach in Category S, which is the least relevant as it has high overlapping and no disease progression, is also a classical approach, REMPE. This method involves an eye-specific deformation model tailored specifically to CF images. It should be noted that all other methods shown in Table \ref{tab:sota} or Figure \ref{fig:sota} use a homographic transformation.

In terms of Category A, the best methods are SuperRetina \cite{eccv20} and SuperRetina with reverse knowledge-distillation \cite{kdsr}. Both are supervised methods, requiring labeled  keypoints for both detection and description. Additionally, these methods have several drawbacks that are not present in other similar methods, like image pre-processing and a double inference step, which significantly increase computational cost and reduce flexibility. Contrary to these approaches, our method is fully unsupervised, a significant advantage in terms of flexibility and adaptability. We can make use of the many CF datasets without a suitable ground truth, making adaptation to other devices much easier. Specifically, compared to SuperRetina and its derivatives, our method produces registrations in a single inference step using lighter models and more efficient descriptors (fewer features), especially when compared to the reverse knowledge distillation variant of SuperRetina.

\begin{table}[]
\centering
\resizebox{0.65\textwidth}{!}{%
\begin{tabular}{@{}lcccccc@{}}
\toprule
\multicolumn{1}{l}{FIRE Dataset}            & FIRE           & A              & P               & S             & Avg.            & W. Avg.        \\ \midrule

\multicolumn{1}{l}{\textit{Classical:}}    &                &                &                 &               &                 &                \\
VOTUS \cite{votus}                  & \textbf{0.812} & 0.681          & \textbf{0.672} & 0.934          & \textbf{0.762} & \textbf{0.811} \\
REMPE \cite{rempe}                           & 0.773 & 0.66           & 0.542 & \textbf{0.958} & 0.72  & 0.774 \\
\midrule
\multicolumn{1}{l}{\textit{Supervised:}}    &                &                &                 &               &                 &                \\
SuperRetina \cite{eccv20}                                & -              & \textbf{0.783} & 0.542*          & 0.94          & 0.755*          & 0.780*         \\
KD-SuperRetina \cite{kdsr}                              & \textbf{-}     & \textbf{0.783} & 0.558* & 0.942         & 0.761* & 0.785*         \\
ConKeD \cite{rivas3}                                     & 0.758          & 0.749          & 0.489           & 0.945         & 0.728           & 0.758          \\
ConKeD++ \cite{rivas4}                                   & 0.76           & 0.766          & 0.503           & 0.945         & 0.738           & 0.765          \\
\textit{Ours w/ CB kpts}                                  & 0.769          & 0.757          & 0.513           & 0.948         & 0.739           & 0.769          \\
Rivas-Villar \cite{rivas}                     & 0.657          & 0.660          & 0.293           & 0.908         & 0.620           & 0.552          \\ \midrule
\multicolumn{1}{l}{\textit{Unsupervised:}}  &                &                &                 &               &                 &                \\
Retina-R2D2 \cite{rivas2}                               & 0.695          & 0.726          & 0.352           & 0.925         & 0.645           & 0.575          \\
\textit{Ours w/ 500 kpts.}              & 0.721          & 0.709          & 0.441           & 0.916         & 0.688           & 0.721          \\
\textit{Ours w/ All kpts.} & 0.768 & 0.745 & 0.520 & 0.943 & 0.736 & 0.768

\\ \midrule
\multicolumn{1}{l}{\textit{Detector-less:}} &                &                &                 &               &                 &                \\
GeoFormer  \cite{geoformer}                                & -              & 0.76           & 0.559*          & 0.944         & 0.754*          & 0.784*         \\
LoFTR  \cite{geoformer}                                      & -              & 0.711          & 0.359*          & 0.92          & 0.663*          & 0.693*         \\
ASpanFormer  \cite{geoformer}                                 & -              & 0.703          & 0.495*          & 0.921         & 0.706*          & 0.742*         \\ \bottomrule
\end{tabular}%
}
\vspace{1em}
\caption{Comparison of our approach with the best state-of-the-art methods. The results are measured in the standard Registration Score AUC. * Indicates that the results are obtained with one less image. The best results for each category are highlighted in bold.}
\label{tab:sota}
\end{table}

Additionally, we also included in Table \ref{tab:sota} an evaluation of our method with all the detectable keypoints. This offers an evaluation that is more similar to detector-less methods which employ a pipeline that uses the whole image. Using all the possible keypoints, we obtain results in line with the best deep-learning-based methods. This version of our approach detects, approximately, 2250 keypoints, which limits efficiency. However, our detector and descriptor method still produces far fewer keypoints than those that would be employed by the detector-less methods.
In this family, GeoFormer achieves the best results by adding RANSAC to the LoFTR pipeline. While this leverages both similarity and geometric constraints, it incurs higher computational costs than the more efficient yet worse-performing LoFTR.

\subsection{Limitations}

The proposed method exhibits specific limitations inherent to its design choices. First, the reliance on a homographic transformation provides fewer degrees of freedom compared to deformation models used by the leading classical approaches, like VOTUS. Consequently, our approach does not match the performance of this method in scenarios requiring large deformation modeling, such as Category P of the FIRE dataset.  Furthermore, the keypoint allocation strategy does not explicitly filter non-overlapping zones (most relevant in Category P of FIRE). This inefficiency consumes the keypoint budget on features that cannot be matched, increasing computational cost without performance gain.  It should be noted that this is a challenge that remains unsolved in the state-of-the-art for keypoint-based and detector-less approaches.

Regarding feature extraction, while both AP and SS maps represent several desirable properties of the keypoints, they do not impose explicit constraints for discriminability against immediate neighborhoods or from repetitive patterns or textures.

Finally, the use of standard NMS can lead to a uniform spatial distribution of keypoints. A density-aware suppression mechanism could improve results and efficiency, especially in keypoint-constrained scenarios, but this benefit comes at the cost of additional parameters to tune.

\section{Conclusions}

In this work, we introduce a novel method to detect and describe keypoints in CF images without relying on any ground truth. While state-of-the-art methods typically use labeled keypoints to train a supervised detector and its corresponding descriptor, our approach inverts this paradigm.
Our method is based on the idea that locations with high-performance descriptors serve as keypoints. To achieve this, we propose an unsupervised descriptor training method that neither requires labeled data nor depends on any detector. Then, by estimating the performance of this descriptor network through heatmaps, we can subsequently train a detector network. This detector network predicts heatmaps directly from the input image, resulting in an unsupervised keypoint detector.

We conducted a comprehensive evaluation across multiple CF datasets with different characteristics. Our results reveal that our novel unsupervised descriptor outperforms state-of-the-art supervised alternatives, even in unfavorable conditions. Similarly, our unsupervised detector improves the results of existing unsupervised detectors. Notably, our complete registration pipeline achieves performance on par with the best current methods, despite being unsupervised. This represents a significant step forward in unsupervised methods, as, especially in the medical domain, labeled images are scarce. 
Beyond CF images, the nature of our method allows direct generalization and application in other image modalities without any modification.

\section*{Acknowledgements}
This work was supported by the Instituto de Salud Carlos III (\mbox{ISCIII}), Government of Spain [grant number \mbox{FORT23/00010} (Programa \mbox{FORTALECE})]; \mbox{MICIU}/\mbox{AEI}/\mbox{10.13039/501100011033} and ``\mbox{ERDF} A way of making Europe'' [grant numbers \mbox{PID2023-148913OB-I00} and \mbox{PID2024-161024OB-I00}]; and the Consellería de Educación, Universidade e Formación Profesional, Xunta de Galicia, Grupos de Referencia Competitiva [grant number \mbox{ED431C 2024/33}], pre-doctoral grant number \mbox{ED481A 2021/147}, and post-doctoral fellowship number \mbox{ED481B-2022-025}. Funding for open access charge: Universidade da Coruña/CISUG.
\bibliographystyle{IEEEtran}
\bibliography{references}

\end{document}